\documentclass[11pt,a4paper]{article}
\usepackage[hyperref]{emnlp2020}
\usepackage[american]{babel}
\usepackage{times}
\usepackage[T1]{fontenc}
\usepackage{amssymb}
\usepackage{microtype}
\usepackage{url}
\usepackage{multirow} 
\usepackage{array}

\aclfinalcopy 

\usepackage{xspace}

\usepackage{enumitem}

\RequirePackage{xcolor}
\definecolor{lightgray}{gray}{0.80}
\definecolor{mediumgray}{gray}{0.60}
\definecolor{darkgray}{gray}{0.40}

\usepackage[pdftex]{graphicx}
\usepackage{booktabs}
\DeclareGraphicsExtensions{.pdf,.ai,.jpg,.png}
\setkeys{Gin}{pagebox=artbox}
\graphicspath{{nlpandcss20-bias-detection-paper-figures/}}

\newcommand{\bsfigure}[3][]{
  \begin{figure}[t]
  \centering
  \includegraphics[#1]{#2}
  \caption{#3}\label{#2}%
  \end{figure}}

\RequirePackage{color}
\RequirePackage{soul}
\setstcolor{blue}

\newsavebox\bscombox
\newcommand{\bscom}[3][]{%
  \sbox{\bscombox}{\fontsize{8}{9}\selectfont#1#2#3}
  \noindent
  \st{#2}{\selectfont
    \color{blue}#3\ifx\\#1\\\else{\fontsize{8}{9}\selectfont\color{violet}[#1]}\fi
    }
  }

\begin{document}

\title{Analyzing Political Bias and Unfairness in News Articles \\[0.2ex] at Different Levels of Granularity}

\author{
Wei-Fan Chen \\
Paderborn University \\
Department of Computer Science \\
{\tt cwf@mail.upb.de} \\\And
Khalid Al-Khatib \\
Bauhaus-Universit\"at Weimar \\
Faculty of Media, Webis Group \\
{\tt khalid.alkhatib@uni-weimar.de } \\\AND
Henning Wachsmuth \\
Paderborn University \\
Department of Computer Science \\
{\tt henningw@upb.de} \\\And
Benno Stein\\
Bauhaus-Universit\"at Weimar \\
Faculty of Media, Webis Group \\
{\tt benno.stein@uni-weimar.de}
}

\date{}

\setlength\titlebox{40ex}
\maketitle

\begin{abstract}
Media organizations bear great reponsibility because of their considerable influence on shaping beliefs and positions of our society. Any form of media can contain overly biased content, e.g., by reporting on political events in a selective or incomplete manner. A relevant question hence is whether and how such form of imbalanced news coverage can be exposed. The research presented in this paper addresses not only the automatic detection of bias but goes one step further in that it explores how political bias and unfairness are manifested linguistically. In this regard we utilize a new corpus of 6964 news articles with labels derived from {\em adfontesmedia.com} and develop a neural model for bias assessment. By analyzing this model on article excerpts, we find insightful bias patterns at different levels of text granularity, from single words to the whole article discourse.
\end{abstract}

\section{Introduction}

Reporting news in a politically unbiased and fair manner is a key component of journalism ethics and standards. ``Politically unbiased'' means to report on an event without taking a \emph{political position, characterization, or terminology}, and ``fair'', in this context, means to focus on original facts rather than on \emph{analyses or opinion statements based on false premises}.%
\footnote{See \url{http://www.adfontesmedia.com} for a detailed characterization of political bias and unfairness.}
Although it is known that biased and unfair news do exist in the media, spreading misleading information and propaganda \cite{groseclose:2005}, people are not always aware of reading biased content. 

When fighting bias and unfairness, only the understanding of how these emerge (and how to avoid them) can allow media organizations to maintain credibility and can enable readers to choose what to consume and what not. In pursuit of this goal, we recently studied how sentence-level bias in an article affects the political bias of the whole article \cite{chen:2020}. Also other researchers have proposed approaches to automatic bias detection (see Section~\ref{sec:relatedwork} for details). However, the existing approaches lack an analysis of what makes up bias, and how it exposes in different granularity levels, from single words, to sentences and paragraphs, to the entire discourse.

To close this gap, we analyze political bias and unfairness in this paper within three steps:
\begin{enumerate}
\item
We develop an automatic approach to detect bias and unfairness in news articles.
\item
We study the bias distribution along different text granularity levels.
\item
We explore various sequential patterns of how bias and unfairness become manifest in text. 
\end{enumerate}

We utilize two well-known websites that address media bias, {\em allsides.com} and {\em adfontesmedia.com}, in order to create a corpus with 6964 news articles, each of which is labeled for its topic, political bias, and unfairness. Based on this corpus, we devise a recurrent neural network architecture to learn classification knowledge for bias detection. We choose this network class because of its proven ability to capture semantic information at multiple levels: taking the model output for whole texts, we conduct an in-depth reverse feature analysis to explore media bias at the word, the sentence, the paragraph, and the discourse level. At the {\em word level}, we correlate the most biased sentences with LIWC categories \cite{pennebaker:2015}. At the {\em sentence and paragraph level}, we reveal what parts of an article are typically most politically biased and unfair. At the {\em discourse level}, we reveal common sequential media bias patterns.

The results show that our model can utilize high-level semantic features of bias, and it confirms that they are manifested in larger granularity levels, i.e., on the paragraph level and the discourse level. At the word level, we find some LIWC categories to be particularly correlated with political bias and unfairness, such as {\em negative emotion}, {\em focus present}, and {\em percept}. At levels of larger granularity, we observe that the last part of an article usually tends to be most biased and unfairest.

\section{Related Work}
\label{sec:relatedwork}

Computational approaches to media bias focus on factuality \cite{baly:2018}, political ideology \cite{iyyer:2014}, and information quality \cite{rashkin:2017}. Bias detection has been done at different granularity levels: single sentences \cite{bhatia:2018}, articles \cite{kulkarni:2018}, and media sources \cite{baly:2019}. Recently, the authors of this paper studied how the two granularity levels ``sentence'' and ``discourse'' affect each other. An outcome of this research are insights and means of how sentence-level bias classifiers can be leveraged to better predict article-level bias \cite{chen:2020}.

Given these results, the paper in hand digs deeper by investigating how bias is actually manifested in a text at different granularity levels. Our approach resembles techniques where the attention mechanism in a model is used to output weights (which indicate feature importance) for each text segment \cite{bahdanau:2014}. \citet{zhou:2016}, for instance, use word-level attention to focus on sentiment words, while \citet{ji:2017} use sentence-level attention to select valid sentences for entity relation extraction. Related to media bias, \citet{kulkarni:2018} show that the learned attention can focus on the biased sentences when detecting political ideology.

Regarding ``attention'' at multi-level granularity, \citet{yang:2016} propose a hierarchical attention network for document classification that is used at both word and sentence level. Yet, problems when using attention for such analyses are that the analysis unit (be it a word or a sentence) has to be defined before the training process, and that the set of classifiers which can output attention is limited. By contrast, our unsupervised reverse feature analysis can be used with any classifier and at an arbitrary semantic level, even after the training process.

\section{Media Bias Corpus}
\label{sec:data}

Although bias detection is viewed as an important task, we could not find any reasonably sized labeled corpus that suits our study, which is why we decided to built a new one. We started with the corpus of \citet{chen:2018} in order to obtain news articles with topics and political bias labels. We extended the corpus by crawling the latest articles from {\em allsides.com} and by adding the fairness labels provided by {\em adfontesmedia.com}. The new corpus is available at \url{https://github.com/webis-de/NLPCSS-20}.

\paragraph{Allsides.com}

This website is a news aggregator that collects news articles about American politics, starting from June 1st, 2012. Each event comes with articles representing one of the three political camps: left, center, and right. \citet{chen:2018} crawled the website to extract 6447 articles. For our study, we extended their corpus by integrating all articles until March 15th, 2019, resulting in a total of 7775 articles. In addition to the political bias labels, we crawled the topic tags of each article.

\paragraph{Adfontesmedia.com}

Since allsides.com focuses on political bias, we exploit adfontesmedia.com as another source for additional bias types. This portal maintains a ``bias scale'' quantifying the media bias of a broad set of US news portals. The bias assessments stem from media experts who annotate each portal with bias and fairness labels. \citet{bentley:2019} show that the portals' labels are highly correlated to findings from social scientists.

\paragraph{A New Media Bias Corpus}

Based on the labels from adfontesmedia.com, we define three media bias types for news portals: 
\begin{enumerate}
\item
{\bfseries Political Bias.} A portal is {\em neutral} if it is labeled as ``skew left/right'' or ``neutral''. It is {\em politically biased} if it is labeled with ``most extreme left/right'' or ``hyperpartisan left/right''.
\item
{\bfseries Unfairness.} A portal is considered {\em fair} if it is labeled as ``original fact reporting'', ``fact reporting'', ``mix of fact reporting and analysis'', ``analysis'', or ``opinion''. The portal is considered {\em unfair} if it is labeled as ``selective story'', ``propaganda'', or ``fabricated info''.
\item
{\bfseries Non-Objectivity.} A portal is considered {\em objective} if it is politically unbiased and fair. Otherwise, it is considered as {\em non-objective}.
\end{enumerate}

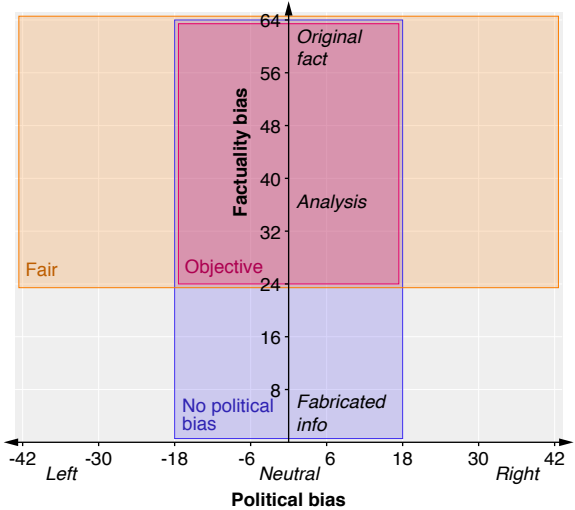
\bsfigure{bias-chart}{The bias chart from {\em adfontesmedia.com} (visually adjusted). The three rectangles represent the positive counterparts of the regions of the three bias types (\emph{political bias}, \emph{unfairness}, and \emph{non-objectivity}).}

Figure~\ref{bias-chart} gives an overview of the labels; it is based on the bias chart at adfontesmedia: The political bias focuses on the x-axis of the chart, while the unfairness focuses on the y-axis of the chart. The non-objectivity is the combination of the two kinds of bias.

\begin{table}[t]
\centering
\small

\renewcommand{\arraystretch}{1.1}
\begin{tabular}{@{}lrlr@{}}

\toprule
\multicolumn{2}{c}{\bf Portal}  & \multicolumn{2}{c}{\bf Topic}\\
\cmidrule(l@{2pt}r@{2pt}){1-2}
\cmidrule(l@{2pt}r@{2pt}){3-4}
Name&Count&Name&Count\\
\midrule
CNN            & 1021 & presidential election & 914\\
Fox News       & 1002 & politics              & 525\\
New York Times & 781  & White House           & 515\\
\addlinespace[3px]
... & &... &\\
\addlinespace[3px]
NPR News       & 1    & domestic policy       & 2\\
The Nation     & 1    & EPA                   & 1\\
Vice           & 1    & women's issues        & 1\\

\bottomrule
\end{tabular}

\caption{The top three and the bottom three portals along with the topics in our corpus.}
\label{table-corpus}
\end{table}

We label the collected articles according to this scheme. Since adfontesmedia.com does not cover all portals from allsides.com, the final corpus contains 41 portals with 6964 articles. The three largest portals are CNN (1021 articles), Fox News (1002 articles), and the New York Times (781 articles). Altogether, we count 111 different topics such as, ``presidential election'' (914 articles), ``politics'' (525 articles), and ``white house'' (515 articles). Table~\ref{table-corpus} lists the top three and the bottom three portals along with topics in our corpus.

\section{Media Bias Analysis}
\label{sec:exp_ana}

This section presents our approach to study bias and unfairness at different levels of text granularity. At first, we develop classifiers for political bias, unfairness, and non-objectivity. Second, we do a reverse feature analysis study to explore the manifestation of bias at multiple levels of granularity and to identify different bias patterns.

\subsection{Media Bias Classification}

For the detection of bias in a text, we follow \citet{chen:2018} in keeping the word order, in order to be able to capture higher-level semantics. To this end, we develop classifiers with RNN served as the classical model for sequential inputs, where a cell is a GRU with a recurrent state size of 32. On top of the final hidden vector of GRUs is a prediction layer whose activation function is a softmax and the size is 2. We use the pre-trained word embedding of GloVe \cite{pennington:2014} with a word embedding dimension of~50, the optimizer Adam, and a learning rate of~0.001. We train classifiers until no improvement in the development set is observed anymore; all classifiers are of the same structure and have the same hyperparameters.

To minimize the mnemonic information induced by the article topic, we split the dataset controlling the topics as an independent variable: we group the articles by their topic and select some groups to be in the test set, some to be in the development set, and the rest to be in the training set. We ensure that either the development set or the test set has at least 10\% of the articles in the whole dataset, obtaining 5394 articles in the training set, 755 articles in the development set, and 815 articles in the test set. 

Table~\ref{table-label} shows the distribution of the labels in the corpus. To avoid the exploitation of portal-specific features, each article is thoroughly checked and all information regarding the portal it was taken from (e.g., ``CNN's Clare Foran and Phil Mattingly contributed to this report'') is removed.

\begin{table}[t]

\small
\renewcommand{\arraystretch}{1.1}
\centering
\setlength{\tabcolsep}{8pt}
\begin{tabular*}{\linewidth}{@{}rrrr@{}}
\toprule
{\bf Media Bias} & {\bf Training} & {\bf Development} & {\bf Test}\\
\midrule
Political Bias  & 39.25\% & 40.00\% & 42.82\% \\
Unfairness      & 18.59\% & 17.48\% & 18.16\% \\
Non-objectivity & 39.84\% & 40.66\% & 43.31\% \\
\bottomrule
\end{tabular*}
\caption{The percentage of articles with each considered media bias type in the three datasets of our corpus.}
\label{table-label}
\end{table}

\subsection{Reverse Feature Analysis}

In addition to the effectiveness of our bias classification, we want to assess the kind of features that are learned. Formally, we use the developed classifiers to output the predicted bias probability $p_{art}$ of the test articles, i.e., the probability of being \emph{politically biased}, \emph{unfair}, or \emph{non-objective}. We iteratively remove text segments from the article and use the classifier to again predict the media bias probability $p_{art-i}$, where $i$ denotes the index of the text segment in the article. The media bias strength of a text segment $t_i$ is estimated as $p_{art} - p_{art-i}$. If a text segment is relevant for prediction, we expect to see a significant decrease from $p_{art}$ to $p_{art-i}$. 

Based on this estimation of media bias strength, we design three experiments to analyze and interpret the classifiers' predictions at the following levels of text granularity:

\paragraph{Word level (LIWC correlations)}

Related research suggests that media bias is manifested at a larger granularity level, including the paragraph level \cite{chen:2018} and the clause level \cite{iyyer:2014}. To validate this, we use the LIWC categories to check the word level bias, because they have been used in \citet{iyyer:2014} to sample a set of sentences that may contain ideology bias. 

In detail, for each sentence $s_i$, we compute its LIWC score of the category $j$ as $|\{w_{i,k} \in c_j, k \in K \}| \,/\, |\{w_{i,k}, k \in K \}|$, where $c_j$ denotes the words in LIWC category $j$, $K$ denotes the bag-of-words in $s_i$, and $w_{i,k}$ denotes the $k$-th word in $K$. The Pearson correlation coefficient is used to measure the correlation between LIWC categories and media bias strength.

\paragraph{Sentence and paragraph-level (locations of media bias)} 

Here we analyze the distribution of the media bias strength in the sentences and paragraphs (approximated as three continuous sentences). Basically, these values indicate which segment of a text mostly contains media bias.

\paragraph{Discourse level (media bias patterns)}

Here we analyze the patterns of the media bias strength across the different parts of an article's discourse. In particular, we split an article into four equally-sized parts and compute the average media bias strength of the sentences for each part. The splitting is comparable to the so-called ``inverted pyramid'' structure in journalism, where a news article starting by the summary, important details, general and background info~\cite{po:2003}.

\section{Results and Discussion}
\label{sec:discussion}

\begin{table}[t]
\centering
\small

\renewcommand{\arraystretch}{1.1}
\setlength{\tabcolsep}{2.5pt}
\begin{tabular}{lrrr}

\toprule
& \bfseries Political Bias & \bfseries Unfairness & \bfseries Non-objectivity  \\
\midrule
Majority& 36.38\% & 45.01\% & 36.18\% \\
\addlinespace
RNN & 75.60\% & 83.42\% & 75.42\% \\
\hspace{5px} - Biased& 69.41\%& 72.09\% & 69.57\%\\
\hspace{5px} - Unbiased& 81.80\% & 94.75\% & 81.13\%\\
\bottomrule
\end{tabular}
\caption{The F$_1$ scores of RNN, majority baseline, and by-class performance of the three bias types.}
\label{table-classification}
\end{table}

In the following, we report and discuss the results of the experiments described in Section~\ref{sec:exp_ana}.

\subsection{Media Bias Classification}

First, we look at the automatic classification of media bias. Table~\ref{table-classification} summarizes the performance of the developed RNNs in the three media bias classes. All classifiers outperform the majority baseline, achieving 75.60\% for political bias, 83.42\% for unfairness, and 75.42\% for non-objectivity. Such a performance demonstrates the capability of the classifiers to detect topic-independent media bias features. 

Looking closely at individual bias classes, we find that the RNN is good at predicting the absence of bias rather than bias. We interpret this because of the uneven distribution of the classes, especially in the unfairness (see Table~\ref{table-label}).

\subsection{Reverse Feature Analysis}

Although the classifiers achieved good results, they are certainly not perfect. To account for the prediction errors, the following analysis uses only the predicted probabilities of those articles where the RNNs predicted the label correctly.

\paragraph{LIWC Correlations}

According to the Pearson correlations, most of the LIWC categories are not correlated with a high coefficient (neither in a positive or a negative way). However, the highest correlated categories are different among the three types of media bias. The categories that have the highest correlation with political bias are \emph{negative emotion, anger}, and \emph{affect}. This shows that politically biased articles tend to use emotional and opinionated words such as ``disappoint'', ``trust'', and ``angry''. For unfair articles, we see a higher correlation in \emph{focus present}. Examples in this category are ``admit'', ``become'', and ``determine''. For non-objective articles, the bias is related to \emph{percept} words such as ``feel'', ``gloom'', and ``depict''.

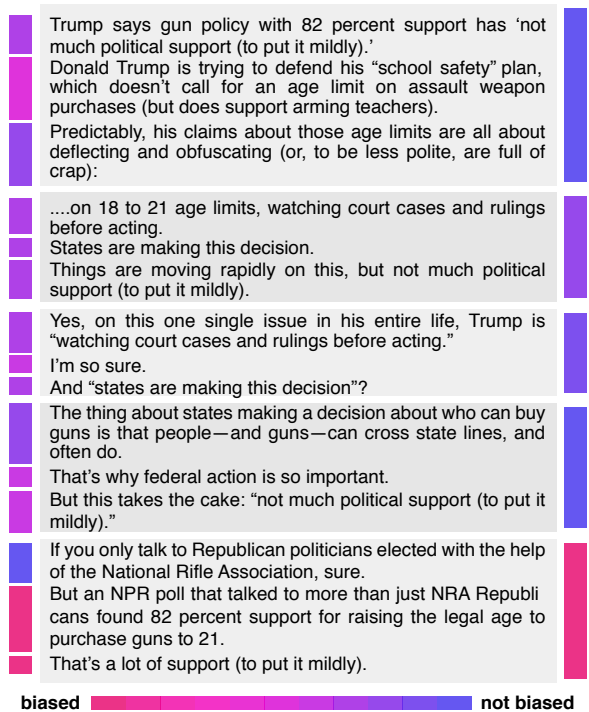
\bsfigure{bias_strength_single2}{The media bias strength on sentence (left) and paragraph (right) level in an excerpt of one news article from the given corpus. Biased text segments are shown in red and unbiased text segments in blue.}

\paragraph{Location of Media Bias}

Figure~\ref{bias_strength_single2} visualizes the estimated media bias strength at the sentence and paragraph levels. As an example, we choose an article from Daily Kos, which is labeled as politically biased. In this article, we see a strong tendency to criticize Trump's claim, especially at the end of the article. At the paragraph level, our strength analysis of media bias successfully identifies the last paragraph as the most biased text segment. While at the sentence level, we see that the last two sentences are most biased in the last paragraph. The second sentence seems to be a bit biased, perhaps because of the word usage of ``trying to defend''. However, we see that the analysis fails to identify the third sentence as politically biased. Still, given that the sentence or paragraph level analysis is fully unsupervised, the reverse feature analysis seems to perform quite well.

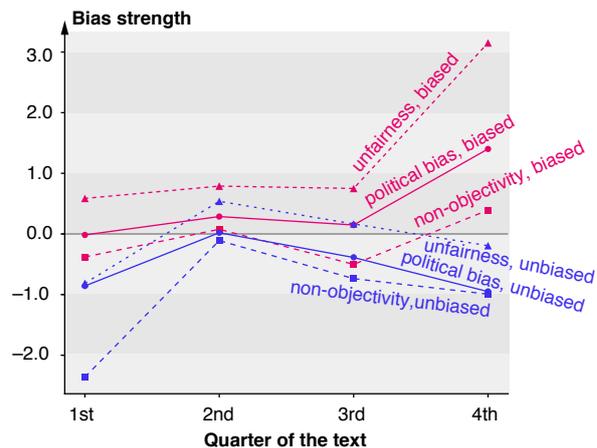
\bsfigure{pattern}{Patterns of the types of media bias as well as biased and unbiased text. Values are normalized to have a mean of zero and a standard deviation of one. Positive values indicate a stronger bias and negative values indicate that text has a lower bias or is unbiased.}

\paragraph{Media Bias Patterns}

Figure~\ref{pattern} shows the identified media sequential patterns for the three bias types. We can notice that the media bias strengths for all articles in the second quarter are somewhat close. This is, in our opinion, because the second quarter of news articles usually contains some background information, which does not tend to be biased. We also see that all biased articles start with a neutral tone (close to mean) in the beginning and then emphasize the bias in the latter parts. Among the three media bias types, unfairness has the highest bias strength. Observing our corpus, we find that one typical way to be unfair is to report selected facts in a favor of some entity, which leads to completely different word usage. On the other hand, for political bias, describing facts with positive or negative expressions is a common indicator of bias there. Such a difference might be the reason for why the classifiers can better discover the unfair texts.

\section{Conclusion}
\label{sec:conclusion}

This paper has studied political bias and unfairness in news articles. We have trained sequential models for bias detection and have applied a reverse feature analysis to demonstrate that it is possible to reveal at what granularity level and how sequential patterns media bias is manifested. Specifically, we find that the last quarter seems to be the most biased part. A significant ``by-product'' of our research is a new corpus for bias analysis. And we believe this corpus can help, for example, investigate how journalists convey bias in a news article.

In the future, we will study how to utilize these results to detect bias or to rewrite the articles to remove and change bias. For example, based on the findings of this paper, the bias remover should pay more attention to the later paragraphs.

\section*{Acknowledgments}

This work was partially supported by the German Research Foundation (DFG) within the Collaborative Research Center ``On-The-Fly Computing'' (SFB~901/3) under the project number~160364472.

\bibliography{nlpcss20-bias-detection-lit}
\bibliographystyle{acl_natbib}

\end{document}